\documentclass[a4paper,UKenglish,cleveref, autoref, thm-restate]{lipics-v2021}

\hideLIPIcs  

\usepackage{newfloat}
\usepackage{listings}

\lstdefinelanguage{stripspddl}
{
  sensitive=false,    
  morecomment=[l]{;}, 
  alsoletter={\#,:},   
  morekeywords={
    not, \#blackactions, \#whiteactions,
    :action,:parameters,:precondition,:effect,
    \#init,\#blackgoal,\#whitegoal,\#boardsize, \#depth, \#blackturn
  }
}

\usepackage{tikz}
\usetikzlibrary{arrows,snakes}
\usetikzlibrary{shapes.geometric}
\usetikzlibrary[calc,shapes,patterns,positioning,decorations.pathmorphing,decorations.pathreplacing]
\usepackage{pgfplots}
\DeclareUnicodeCharacter{2212}{−}
\usepgfplotslibrary{groupplots,dateplot}
\pgfplotsset{compat=newest}

\definecolor{light-gray}{gray}{0.92}

\newcommand{\Xmax}{5}
\newcommand{\Ymax}{4}
\newcommand{\step}{1}

\newcommand{\halfstep}{\step/2}

\newcommand{\havcoordinate}[4]{
  \foreach \i in {#1,...,#2} {
    \foreach \j in {#3,...,#4} {
      \coordinate (\i-\j) at ({(\i)-\halfstep},{\Ymax-(\j-\halfstep)});
    }
  }
}

\newcommand{\drawgrid}[4]{
\draw[step=\step cm,black,thin] (0,0) grid (#2,#4);

    \foreach \x in {#1, ..., #2} {%
      \node at (\x - \halfstep,#4+\halfstep) {\x};
      }
    \foreach \y in {#3, ..., #4} {%
      \node at (-0.2,#4-\y+\halfstep) {\y};
      }
	}

\newcommand{\havecircle}[3]{\node[circle,draw,fill=#3,minimum size=0.9*\step cm] (c) at  (#1-#2) {};}

\tikzstyle{startstop} = [rectangle, rounded corners, text centered, draw=black]
\tikzstyle{io} = [trapezium, trapezium left angle=70, trapezium right angle=110, text centered,text width = 2cm, draw=black]
\tikzstyle{decision} = [diamond, text centered, draw=black]
\tikzstyle{arrow} = [thick,->,>=stealth]
\tikzstyle{process} = [rectangle, text centered,text width = 2cm, draw=black]

\usepackage{subcaption}

\title{Concise QBF Encodings for Games on a Grid (extended version)}

\author{Irfansha Shaik}{Department of Computer Science, Aarhus University, Denmark}{irfansha@cs.au.dk}{0000-0002-7404-348X}{} 
\author{Jaco van de Pol}{Department of Computer Science, Aarhus University, Denmark}{jaco@cs.au.dk}{0000-0003-4305-0625}{} 

\authorrunning{Irfansha Shaik and Jaco van de Pol} 

\Copyright{Irfansha Shaik and Jaco van de Pol} 



\relatedversion{} 

\nolinenumbers 

\EventEditors{John Q. Open and Joan R. Access}
\EventNoEds{2}
\EventLongTitle{26th International Conference on Theory and Applications of Satisfiability Testing (SAT 2023)}
\EventShortTitle{SAT 2023}
\EventAcronym{SAT}
\EventYear{2023}
\EventDate{July 04-08}
\EventLocation{Alghero, Italy}
\EventLogo{}
\SeriesVolume{42}
\ArticleNo{23}

\usepackage{amsmath}
\usepackage{amssymb}
\usepackage{booktabs,multirow}
\usepackage{todonotes}

\usepackage{mathtools}
\usepackage{wrapfig}
\usepackage{amsthm}

\DeclareMathOperator{\move}{M} 
\DeclareMathOperator{\I}{I} 

\DeclareMathOperator{\x}{x} 

\DeclareMathOperator{\Conditions}{{\cal C}} 

\DeclareMathOperator{\action}{A} 
\DeclareMathOperator{\X}{X} 
\DeclareMathOperator{\Y}{Y} 
\DeclareMathOperator{\gamestop}{gs} 
\DeclareMathOperator{\legalbound}{lb} 
\DeclareMathOperator{\preflags}{P} 

\DeclareMathOperator{\W}{W} 
\DeclareMathOperator{\B}{B} 

\DeclareMathOperator{\white}{w} 
\DeclareMathOperator{\open}{o} 

\DeclareMathOperator{\whitesymbol}{white} 
\DeclareMathOperator{\blacksymbol}{black} 
\DeclareMathOperator{\opensymbol}{open} 

\DeclareMathOperator{\pre}{pre} 
\DeclareMathOperator{\eff}{eff} 
\DeclareMathOperator{\all}{all} 

\DeclareMathOperator{\sympos}{S} 

\DeclareMathOperator{\allindices}{Ind} 

\DeclareMathOperator{\lessthan}{\textsc{LT}} 
\DeclareMathOperator{\adder}{\textsc{Add}} 
\DeclareMathOperator{\subtractor}{\textsc{Sub}} 
\DeclareMathOperator{\compute}{\textsc{AddSub}} 
\DeclareMathOperator{\boundcircuit}{\textsc{BDcir}} 
\DeclareMathOperator{\equality}{\textsc{EQ}} 

\DeclareMathOperator{\vars}{V} 
\DeclareMathOperator{\symbimp}{RI} 

\DeclareMathOperator{\subcon}{SC} 

\DeclareMathOperator{\goal}{G} 

\DeclareMathOperator{\bounds}{BD} 

\DeclareMathOperator{\ymax}{ymax} 
\DeclareMathOperator{\xmax}{xmax} 

\DeclareMathOperator{\ymin}{ymin} 
\DeclareMathOperator{\xmin}{xmin} 

\DeclareMathOperator{\Icon}{\mathcal{I}} 
\DeclareMathOperator{\Gcon}{\mathcal{G}} 

\DeclareMathOperator{\win}{won} 
\DeclareMathOperator{\WIN}{\mathcal{W}} 

\DeclarePairedDelimiter\ceil{\lceil}{\rceil}

\begin{document}

\maketitle

\begin{abstract}
Encoding 2-player games in QBF correctly and efficiently is challenging and error-prone.
To enable concise specifications and uniform encodings of games played on grid boards, 
like Tic-Tac-Toe, Connect-4, Domineering, Pursuer-Evader and Breakthrough,
we introduce BDDL -- \emph{Board-game Domain Definition Language},
inspired by the success of PDDL in the planning domain.

We provide an efficient translation from BDDL into QBF, encoding the existence of a winning strategy of bounded depth.
Our lifted encoding treats board positions symbolically and allows concise definitions of conditions, effects and winning configurations,
relative to symbolic board positions. 
The size of the encoding grows linearly in the input model and the considered depth.

To show the feasibility of such a generic approach, we use QBF solvers to compute the
critical depths of winning strategies for instances of several known games. 
For several games, our work provides the first QBF encoding.
Unlike plan validation in SAT-based planning, validating QBF-based winning strategies 
is difficult.
We show how to validate winning strategies using QBF certificates and interactive game play.
\end{abstract}

\section{Introduction}
\label{sec:introduction}

The existence of a bounded winning strategy for 2-player games can be encoded elegantly with 
Quantified Boolean Formulas (QBF) \cite{DBLP:series/faia/BeyersdorffJLS21}, where 
the moves of Player 1 and 2 are encoded using existentially and universally quantified variables, respectively.
General QBF solvers have been applied to solve several specific games. 
For instance, the first QBF encoding for Connect-4 \cite{Gent2003EncodingCU} 
was a response to a challenge posed earlier \cite{walsh2003challenges}.
Solving the full Connect-4 game this way was not possible. 
It is a challenge to tune the encoding for pruning the search space.
Another example is the simple chess-like game Evader-Pursuer.
A first QBF encoding \cite{Alur2004SymbolicCT} was improved by an
encoding that guides the solver to prune the search-space \cite{Ansotegui2005AHQ}.
In recent years, there has been progress in encoding positional games like Tic Tac Toe and Hex.
The previous Connect-4 encoding was adapted to positional games
\cite{DBLP:conf/sat/DiptaramaYS16}. Later, the corrective encoding 
was proposed \cite{10.1007/978-3-030-51825-7_31}, which improves pruning by 
correcting illegal white moves instead of using indicator variables. The encoding was
further improved to allow a pairing strategy \cite{BoucherV2021}. A concise,
lifted encoding for positional games was recently introduced in \cite{shaik2023HexArxiv}.


The literature above focused on encoding specific games efficiently, 
which can be challenging and error-prone.
Here we present a uniform translation that can handle a wide range of games, both positional and non-positional games.
The first step is to decouple the modelling of games and their encoding in QBF.
Inspired by the success of PDDL (Planning Domain Definition Language) \cite{DBLP:journals/jair/FoxL03}
and GDL (Game Description Language) \cite{DBLP:journals/aim/GeneserethLP05},
we introduce BDDL (Board-game Domain Definition Language) to specify games played on grid-like boards.
This allows concise specifications of games like Tic-Tac-Toe, Connect-4, Domineering, Pursuer-Evader and BreakThrough.

We propose an efficient encoding of BDDL into QBF, for the existence of a winning strategy of bounded depth. Copying explicit goal and move constraints for each position 
can blow up quickly, even for small boards. Instead, we provide 
the first lifted QBF encoding for \emph{non-positional games}, completely avoiding
grounding of concrete board positions.
A lifted encoding for classical planning was presented in \cite{DBLP:conf/aips/ShaikP22},
and an extension to \emph{positional games} was provided in \cite{shaik2023HexArxiv}.
Here universal symbolic variables were used to specify conditions on moves and goals just once,
for a single, isolated symbolic position.

The problem for \emph{non-positional games} is that the conditions and effects for a move on one position can depend on the state of other positions, for instance when moving or taking pieces. We extend the lifted encoding
to this case, by using the structure of the grid to encode conditions on the neighborhood of a position symbolically.
We use adder, subtractor and comparator circuits to efficiently handle out-of-bounds constraints and illegal moves.
We use existential indicator variables with nested constraints, to avoid unnecessary search space
over illegal white moves.
Finally, we use a single copy of goal configurations, and use universal variables to check
the goal at every time step.

To show the feasibility of such a generic approach, we use QBF solvers to compute optimal winning strategies for small instances of several known games.
We also compare the results to some existing QBF game encodings. Our work provides the first QBF encoding for several games such as BreakThrough, KnightThrough and Domineering.

Validating results when using QBF is nontrivial. Errors can occur both during encoding and solving.
In Section \ref{subsec:correctnessandvalidation}, we present a framework for validating winning strategies generated by our encoding using QBF certificates and interactive game play.
We visualize the state of the board directly from the certificate, which can help with detecting errors.

\section{Preliminaries}
\label{sec:preliminaries}

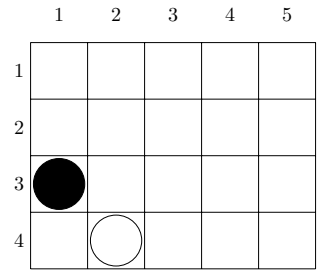
\begin{wrapfigure}{r}{0.3\textwidth}
\centering
\vspace{-4ex}
\scalebox{.75}{\begin{tikzpicture}  
\havcoordinate{1}{\Xmax}{1}{\Ymax}
\drawgrid{1}{\Xmax}{1}{\Ymax}
\havecircle{1}{3}{black}
\havecircle{2}{4}{white}
\end{tikzpicture}}
\caption{HTTT Tic,\\ partially filled 5x4 board}
\label{fig:ticexample}
\end{wrapfigure}

In \emph{2-Player, turn-based Games}, players Black (first player) and White try to reach some goal condition.
In Maker-Maker games, both players try to achieve their own goal; in Maker-Breaker games, the second player wins by stopping the first player.
In this paper, we consider a subset of games that can be played on a \emph{grid}, such as Hex, HTTT, Connect-c, Breakthrough.
For example, Fig.~\ref{fig:ticexample} shows an instance of HTTT Tic, a positional game, where both players try to form a vertical or horizontal line of 3 positions on the grid.
We only consider games with one type of piece, but our work could easily be extended to more complex games with different pieces, like Chess.

\medskip
The \emph{Planning Domain Definition Language} (PDDL)~\cite{DBLP:journals/aim/McDermott00} 
is a standard domain specification language for classical planning problems used in International Planning Competitions (IPC).
A domain file specifies predicates and actions whereas a problem file specifies objects, initial state and goal condition.
PDDL is an action-centered language; the actions essentially specify how the world changes by preconditions and effects.
Actions conditions and effects are represented similar to First-Order-Logic, using parameters to achieve compact descriptions. 

\medskip
\emph{Quantified Boolean Formulas} (QBF)~\cite{DBLP:series/faia/BeyersdorffJLS21} 
extend propositional logic with Boolean quantifiers.
We consider closed QBF formulas in prenex normal form, i.e., $Q_1 x_1 \cdots Q_n x_n (\Phi)$, where
$\Phi$ is a propositional formula with Boolean variables in $\{x_1,\ldots,x_n\}$ and each $Q_i\in\{\forall,\exists\}$.
Every such formula evaluates to true or false. QBF evaluation is a standard PSPACE-complete problem. It is well known that the complexity increases with the number of quantifier alternations.
Several QBF solvers exist, which operate on QBF in either QDIMACS format, where $\Phi$ is essentially a set of CNF
clauses, or in QCIR format \cite{jordan2016noncnf}, where $\Phi$ is provided as a circuit with and- and or-gates and negation.
In Section~\ref{sec:qbfencoding}, we present our encoding of grid-games in QCIR, which can be transformed to QDIMACS using the Tseitin transformation~\cite{Tseitin1983CDPC},
introducing one existential Boolean variable per gate.

\medskip
\emph{QBF Certificates}, proposed by~\cite{DBLP:conf/ijcai/Benedetti05},
are resolution-proofs or Skolem/Herbrand functions which are used for validating the QBF solver result.
A Skolem function for a True instance is essentially a function mapping from universal to existential variables. These certificates can be used to extract winning strategies for 2-player games.
Existing tools for QBF certificate generation include 
sKizzo, using symbolic solving \cite{DBLP:conf/ijcai/Benedetti05}, 
QBFcert, based on resolution proofs \cite{DBLP:conf/sat/NiemetzPLSB12}, 
and FERPModels, which is expansion based \cite{Ferpmodels2022}.
In Section~\ref{subsec:correctnessandvalidation}, we extract and validate winning strategies using
certificate extraction with QBFcert and interactive game play. Our main purpose is to validate
our QBF encoding, rather than the correctness of the QBF solver.

\section{Board-game Domain Definition Language (BDDL)}
\label{sec:domandprobformat}

\subsection{Syntax of BDDL}
\label{subsec:syntax}
Listings \ref{lst:grammar}, \ref{lst:grammarDomain}, \ref{lst:grammarProblem},
 specify the grammar for BDDL conditions, domain files, and problem files.

\begin{lstlisting}[caption={Grammar for BDDL conditions relative to board position $(?x,?y)$.
We will use \texttt{nl} for newline, \texttt{int} for integers and \texttt{str} for
ASCII strings of letters, digits and underscores.
We use \texttt{|} for choice, \texttt{*} for 0 or more repetitions. 
All other symbols are non-terminals or literals.}%
,label={lst:grammar},language=stripspddl,mathescape]
condition $\Coloneqq$ (sub-cond*) nl
sub-cond  $\Coloneqq$ pred(e1,e2) | NOT(pred(e1,e2))
pred      $\Coloneqq$ open | white | black
e1        $\Coloneqq$ ?x + int | ?x - int | ?x | int | xmin | xmax
e2        $\Coloneqq$ ?y + int | ?y - int | ?y | int | ymin | ymax
\end{lstlisting}

We follow the separation of predicates and actions from PDDL. 
For our purpose, the predicates are fixed to $\{\blacksymbol, \whitesymbol,\opensymbol\}$.
Contrary to PDDL, we allow structured expressions in position conditions (see \texttt{(e1,e2)} in Listing \ref{lst:grammar}).
This allows us to specify multiple positions in the grid, relative to an arbitrary
board position $(?x,?y)$.

\begin{definition}
\label{def:condition}
A \emph{condition} is a sequence of sub-conditions $p(e_1,e_2)$ or $\neg p(e_1,e_2)$, where the predicate $p$
and relative coordinates $(e_1,e_2)$ are defined according to the grammar in Listing~\ref{lst:grammar}.
We write $\Conditions$ for the set of all conditions. A condition will be interpreted as a conjunction.
\end{definition}

\paragraph*{Domain Specification}
\label{para:domainformat}

\begin{wrapfigure}[8]{r}{0.5\textwidth}
\vspace{-5ex}
\begin{lstlisting}[caption={Grammar for BDDL domain files.}%
,label={lst:grammarDomain},language=stripspddl,mathescape]
domain $\Coloneqq$ #blackactions nl action* 
          #whiteactions nl action*
action $\Coloneqq$ :action str nl 
          :parameters (?x,?y) nl
          :precondition condition 
          :effect condition
\end{lstlisting}
\end{wrapfigure}

The domain file (Listing~\ref{lst:grammarDomain})
specifies the set of actions that black and white players can play. Each action is defined uniformly over the board
positions by using fixed parameters $(?x,?y)$ as coordinates. Each action is specified by its precondition and effect,
both consisting of a conjunction of positive or negative single position conditions.

In the domain file, absolute int-indices as in $\blacksymbol(1,2)$ \emph{are not allowed} (since the board size is unknown),
but we allow references to the minimal and maximal $(x,y)$-positions.
Listing \ref{lst:positionaldomain} shows an example domain for positional games. 
Section \ref{subsec:domandprobgames} will illustrate more games.

\begin{definition}
\label{def:gamedomain}
Each domain file according to the grammar in Listing~\ref{lst:grammarDomain}
defines a \emph{Game Domain} $(\action_b, \action_w, \pre, \eff)$,
where 
\begin{itemize}
\item $\action_b$ and $\action_w$  are the set of black, resp. white, action symbols specified
\item $\pre:\action_{b} \cup \action_{w}\to \Conditions$ specifies the precondition of each action
\item $\eff:\action_{b} \cup \action_{w}\to \Conditions$ specifies the effect of each action
\end{itemize}
For each action $a \in \action_{b} \cup \action_{w}$, we write
$\all(a) := \pre(a)\cup\eff(a)$ for all its conditions.
\end{definition}

\paragraph*{Problem Instance Specification}
\label{para:problemformat}

\begin{wrapfigure}[8]{r}{0.54\textwidth}
\vspace{-5ex}
\begin{lstlisting}[caption={Grammar for BDDL problem files.}%
,label={lst:grammarProblem},language=stripspddl,mathescape]
problem $\Coloneqq$ size init depth goals
size  $\Coloneqq$ #boardsize nl int int nl
init  $\Coloneqq$ #init nl (pred(int,int)*) nl
depth $\Coloneqq$ #depth nl int nl
goals $\Coloneqq$ #blackgoals nl condition* 
         #whitegoals nl condition*
\end{lstlisting}
\end{wrapfigure}

In the problem file, we specify the rectangular board size $m\times n$, the initial state, the winning conditions
for the black and white player, and we also specify the considered depth $d$ of the game.%
\footnote{We make $d$ part of the problem file for convenience, since we will only consider bounded plays consisting of $d$ moves, but one could study
unbounded plays as well; in both cases, the state space will be finite.}
The initial state is specified by a list of absolute black and white positions.
This list describes a single state, i.e., each position on the board is either black, or white, or open
(in case it is not listed in \#init). Black and White can have multiple alternative winning conditions, each of which is described
by condition on positions. Here we allow both absolute positions (for specific board positions) and relative positions 
(to describe winning patterns uniformly over all board positions).

\begin{definition}
\label{def:gameinstance}
Each problem file according to the grammar in Listing~\ref{lst:grammarProblem} defines
a \emph{Game Instance} $(m,n,\I,\goal_b,\goal_w,d)$, where
\begin{itemize}
\item $(m,n)$ denotes the board size. 
\item $\I \subseteq \{ p(i,j) \mid p\in \{\blacksymbol,\whitesymbol\} \mbox{ and }i,j\in\mathbb{N} \} \subseteq \Conditions$ specifies the initial state.
\item $\goal_b\subseteq\Conditions$ and $\goal_w\subseteq\Conditions$ specify the winning conditions of the black, respectively, white player. 
\item $d$ is an odd number, specifying the considered depth of the play.
\end{itemize}
\end{definition}

For example, Listing~\ref{lst:ticproblem} specifies a $5\times 4$ board, which also determines
$\xmin=\ymin=1$, $\xmax=5$, $\ymax=4$. It specifies two initial board positions:
$(1,3)$ is $\blacksymbol$  and $(2,4)$ is $\whitesymbol$, whereas all other positions are open.
This problem file corresponds to the Tic instance in Figure \ref{fig:ticexample}.
The problem file also specifies that we consider a winning strategy of depth 5.

Finally, the same file specifies two possible winning configuration patterns for black, and two for white.
For instance, white wins as soon as there is a horizontal white line of length 3
 \emph{that fits entirely on the board}. This is specified as
\texttt{white(?x,?y), white(?x+1,?y), white(?x+2,?y)}. Note that the implicit boundary
conditions on $?x$ and $?y$ will be inferred automatically, for ease of specification.

\subsection{Examples: Modelling Some Classical Games}
\label{subsec:domandprobgames}

\paragraph*{Positional Games}
In a positional game, a player can only occupy an open position.
In the domain file, we list black and white actions i.e., a single \emph{occupy} (Listing \ref{lst:positionaldomain}).
We can define goal conditions for games like HTTT and Gomoku implicitly, where we list shapes as disjunction of conjunctions in reference to some existential position.
For example, Listing \ref{lst:ticproblem} is a problem input for the HTTT Tic (this corresponds to Fig.~\ref{fig:ticexample}).
Here the board size is 5x4 and the goal is to form a line of 3 positions either horizontally or vertically.
We can also encode complex goal conditions for games like Hex by listing explicit winning sets of indices.
To encode maker-breaker versions of the games, one can simply drop white goal configurations.

\noindent
\begin{minipage}{0.39\textwidth }
\begin{lstlisting}[caption={Positional games' domain},label={lst:positionaldomain},language=stripspddl,mathescape]
#blackactions
:action occupy
:parameters (?x,?y)
:precondition (open(?x,?y))
:effect (black(?x,?y))
#whiteactions
:action occupy
:parameters (?x,?y)
:precondition (open(?x,?y))
:effect (white(?x,?y))
\end{lstlisting}
\end{minipage}
\begin{minipage}{0.605\textwidth }
\begin{lstlisting}[caption={HTTT Tic problem instance},label={lst:ticproblem},language=stripspddl,mathescape]
#boardsize 5 4
#init      
(black(1,3)white(2,4))
#depth     5
#blackgoals
(black(?x,?y)black(?x,?y+1)black(?x,?y+2))
(black(?x,?y)black(?x+1,?y)black(?x+2,?y))
#whitegoals
(white(?x,?y)white(?x,?y+1)white(?x,?y+2))
(white(?x,?y)white(?x+1,?y)white(?x+2,?y))
\end{lstlisting}
\end{minipage}

\paragraph*{Connect-$c$}

\begin{wrapfigure}{r}{0.38\textwidth}
\vspace{-7.25ex}
\begin{lstlisting}[caption={Connect4: Black}, label={lst:connect4bactions},language=stripspddl,mathescape]
:action occupyOnTop
:parameters (?x,?y)
:precondition 
  (open(?x,?y)
   NOT(open(?x,?y+1)))
:effect (black(?x,?y))
:action occupyBottom
:parameters (?x,?y)
:precondition 
   (open(?x,ymax))
:effect (black(?x,ymax))
\end{lstlisting}
\end{wrapfigure}

\noindent
Connect-$c$ is similar to a positional game, except one can only occupy an open position if the position below it is already occupied.
We can encode this using two actions (Listing \ref{lst:connect4bactions}), \emph{occupyOnTop} and \emph{occupyBottom}.
The last one handles the special case when choosing a position on the bottom row.
Here \emph{ymax} represents the `bottom' row. The actions for white would be similar.
The goal conditions are similar to the Tic problem (Listing \ref{lst:ticproblem}): in Connect-$c$ we would specify a line of 
$c$ positions by its 4 symmetric variants (vertical, horizontal and diagonals).
Connect4 on a $6\times 7$ board is a popular instance.

\paragraph*{Breakthrough}

\begin{wrapfigure}{r}{0.38\textwidth}
\vspace{-10.25ex}
\begin{lstlisting}[caption={Breakthrough (snippets)},
label={lst:breakthrough},language=stripspddl,mathescape]
#blackactions
:action north-east
:parameters (?x,?y)
:precondition 
   (black(?x,?y)
    NOT(black(?x+1,?y-1)))
:effect (open(?x,?y)
         black(?x+1,?y-1))
#blackgoals
(black(?x,ymin))
#whitegoals
(white(?x,ymax))
\end{lstlisting}
\end{wrapfigure}

\noindent
Breakthrough is a non-positional, chess-like game, played with pawns only.
In the Initial board, Black starts with two bottom rows of pawns on $\ymax, \ymax-1$,
 whereas White starts with two top rows of pawns $\ymin, \ymin+1$.
A player can move one step forward or diagonally. On the diagonal steps, it can capture the pieces of the opponent.
The players can only move towards the opposite side of the board.
A player wins if any of its pawns reaches that side.
Model \ref{lst:breakthrough} shows one of the black actions (going north-east with or without capturing), 
and the goal conditions.

\paragraph*{Evader and Pursuer}
In Evader and Pursuer, each player has a single piece starting at different positions of the board.
Black player tries to evade (cannot capture) white player to reach a specific position whereas white player tries to capture black player.
In our model \ref{lst:evaderpursueraction}, Black player can move up to 2 steps vertically or horizontally but only 1 step diagonally.
White player can only move 1 step vertically or horizontally.
In goal conditions (see \ref{lst:evaderpursuergoal}), we simply specify the target position.
Black player wins if it reaches the target position whereas white player wins if it stops black player or reaches the goal first.

\noindent
\begin{minipage}{0.64\textwidth}
\begin{lstlisting}[caption={Evader and Pursuer snippet: 
one of the actions},label={lst:evaderpursueraction},language=stripspddl,mathescape]
#blackactions
:action down-two
:parameters (?x,?y)
:precondition 
  (black(?x,?y) open(?x,?y+1) open(?x,?y+2))
:effect (open(?x,?y)black(?x,?y+2))
\end{lstlisting}
\end{minipage}~~
\begin{minipage}{0.33\textwidth}
\begin{lstlisting}[caption={Goal spec for EP},label={lst:evaderpursuergoal},language=stripspddl,mathescape,escapechar=\%]
#blackgoals
(black(xmin,ymin))
   
#whitegoals
(white(xmin,ymin))
%
\end{lstlisting}
\end{minipage}

\paragraph*{Domineering}
\begin{wrapfigure}[8]{r}{0.45\textwidth}
\vspace{-10.25ex}
\begin{lstlisting}[caption={Domineering (Black only)},
label={lst:domineeringaction},language=stripspddl,mathescape]
#blackactions
:action vertical
:parameters (?x, ?y)
:precondition 
  (open(?x,?y) open(?x,?y+1))
:effect 
  (black(?x,?y) black(?x,?y+1))
  
#blackgoals
#whitegoals
\end{lstlisting}
\end{wrapfigure}

\noindent
In Domineering, the initial board is empty and players take turns to place dominoes.
Black places a domino vertically, covering 2 open positions (illustrated in Listing~\ref{lst:domineeringaction}, top), 
whereas White places dominoes horizontally.
A player wins if the opponent cannot make a move in its turn. This is the default, so we
can just omit the goal conditions (cf.~Listing~\ref{lst:domineeringaction}, bottom).
Note that the empty disjunction corresponds to False.

\medskip
\noindent
While being considerably less general than the Game Description Language (GDL) \cite{DBLP:journals/aim/GeneserethLP05},
one can model many other games in BDDL, e.g., Hex, KnightThrough, CrossPurpose and Konane, etc.
We have chosen a limited setting, in order to provide a uniform encoding in QBF.
To this end, we will now first provide the formal semantics of BDDL.

\subsection{Semantics}
\label{subsec:semantics}
From now on, we assume a fixed Game Domain $(\action_b, \action_w, \pre, \eff)$ and Game Instance $(m,n,\I,\goal_b,\goal_w,d)$,
as introduced in Definitions~\ref{def:gamedomain} and~\ref{def:gameinstance}.

We first define the set of indices $\allindices$ and the implicit bounds induced by using relative positions in the conditions.
For example, the bounds for the second winning condition of white in Listing~\ref{fig:ticexample} is
$[1,3] \times [1,4]$: If $?x\in [1,3]$ and $?y\in [1,4]$, then the positions mentioned in the condition, 
$(?x,?y)$, $(?x+1,?y)$, and $(?x+2,?y)$, refer to actual positions on a 5x4 board.
We assume that all absolute positions in the problem file refer to positions in $\allindices$.

\begin{definition}
\label{def:bounds}
The set of valid indices is defined by $\allindices = [1..m] \times [1..n]$. 
Given a sub-condition $c=p(e_1,e_2)$ or its negation, the \emph{implicit bounds} $\bounds(c)$ are defined as the sub-interval
$[1+\ell_x,m-u_x] \times [1+\ell_y,1-u_y] \subseteq \allindices$, where $\ell_x = a$ if $e_1 = ?x-a$ and $\ell_x = 0$, otherwise.
Similarly, $u_x=a$ if $e_1=?x+a$, $\ell_y=a$ if $e_2=?y-a$, $u_y=a$ if $e_2=?y+a$; otherwise these quantities are 0.

Given a condition $C$, its \emph{implicit bounds} $\bounds(C)$ are defined as the intersection $\bigcap_{c\in C} \bounds(c)$.
\end{definition}

We now define the 2-player game graph induced by a domain and problem file in BDDL.
The nodes of such a graph represent the state space of the game, 
i.e., a game position corresponds to a possible placement of black and white pieces on the board:

\begin{definition}
\label{def:statespace}
We define the set of states $S = \allindices \to \{\blacksymbol,\whitesymbol,\opensymbol\}$. 
We assume that each $(i,j)$ occurs at most once in the init-specification $I$.
Then the initial state $s_0$ is defined such that for all
$(i,j)\in\allindices$, 
$s_0(i,j)=\blacksymbol$ if $\blacksymbol(i,j)\in I$, 
 $s_0(i,j)=\whitesymbol$ if $\whitesymbol(i,j)\in I$, 
  $s_0(i,j)=\opensymbol$, otherwise.
\end{definition}

We can now define when a condition holds in a given game position. For instance,
$\opensymbol(?x,?y) \blacksymbol(?x-1,y) \whitesymbol(?x,y+1)$ holds at tile (2,3)
in the state depicted by Fig.~\ref{fig:ticexample}, because
$\opensymbol(2,3)$, $\blacksymbol(1,3)$ and $\whitesymbol(2,4)$ hold
(fill in (2,3) for $(?x,?y)$).

\begin{definition}
Given a state $s\in S$, a sub-condition $c=p(e_1,e_2)$, and a concrete position
$(i,j)\in\bounds(c)$, we write $e_1[i] \in [1,m]$ for the value of $e_1$ after substituting 
$?x/i$, $\xmin/1$ and $\xmax/m$. Similar for $e_2[j]\in[1,n]$.
Now we define $s, (i,j)\vDash c$ if $s(e_1[i],e_2[j]) = p$. For a condition $C$, and $(i,j)\in\bounds(C)$, 
we define $s, (i,j) \vDash C$ if $s, (i,j) \vDash c$ for all $c\in C$. That is, the conjunction of sub-conditions hold
relative to a single fixed position $(i,j)$.
\end{definition}

We now define the transitions in the game graph.
A player can move from state $s_1$ to $s_2$ if it has an action whose preconditions hold in $s_1$ 
at some valid position $(i,j)$ and whose effects hold in $s_2$ at the same position.

\begin{definition}
\label{def:transition}
Given the set of states $S$ and set of actions $\action$, define for each $a\in\action$ the interval 
$I_a = \bounds(\pre(a))\cap\bounds(\eff(a))$.
We define the transition function
$T(\action) \subseteq S \times S$ by $(s_1,s_2) \in T(\action)$ iff there exists $a \in \action$ and $(i,j)\in I_a$ such that 
$s_1,(i,j)\vDash \pre(a)$ and $s_2,(i,j)\vDash \eff(a)$.
Black's moves are $T(\action_{b})$ and White's moves are $T(\action_{w})$.
\end{definition}

A state is won by a player, if one of its goal conditions holds at some index $(i,j)$.

\begin{definition}
\label{def:blackwinning}
Given a state $s\in S$ and a set of conditions $D$, we write $s\vDash D$ if there exists a condition $C\in D$,
and a position $(i,j)\in\bounds(C)$, such that $s,(i,j)\vDash C$. We define $s\in\win_B$ if $s\vDash \goal_b$
(won by black) and $s\in\win_W$ if $s\vDash \goal_w$.
\end{definition}

In order to simplify the definition of winning strategy (and make our QBF encoding a bit more efficient), we will assume the following
sanity conditions on the game. These conditions hold naturally for all the games that we considered.
The sanity conditions ensure that the winning configuration of a player can only become satisfied
by playing one of its own moves.

\begin{definition}
The 2-player game graph $(S,s_0,T_b, T_w, \win_b, \win_w)$ satisfies the \emph{Sanity Conditions} if:
\begin{itemize}
\item $s_0 \not\in\win_b$ and $s_0\not\in\win_w$ (the initial state is not winning for any of the players)
\item For all $s_1,s_2\in S$, if $T_b(s_1,s_2)$ and $s_2\in \win_w$, then $s_1\in\win_w$.
\item For all $s_1,s_2\in S$, if $T_w(s_1,s_2)$ and $s_2\in \win_b$, then $s_1\in\win_b$.
\end{itemize}
\end{definition}

We are finally in the position to define that Black has a winning strategy of at most $d$ moves.
Here we assume that the players alternate moves, and the first and last move is by Black, so $d$ is always odd. A player who cannot move in its turn has lost.
We define
the sets $\WIN_k$ of states where Black can win in at most $k$ steps, by induction over $k$.

\begin{definition}
\label{def:winningstrategy}
Given a 2-player game graph $(S,s_0,T_b, T_w, \win_b, \win_w)$ that satisfies the sanity conditions, we
define the sets $\WIN_k$, for odd $k$, as follows:
\begin{itemize}
\item $s\in\WIN_1$ if there exists a state $s'$ such that $T_b(s,s')$ and $s'\in\win_b$.
\item $s\in\WIN_{k+2}$ if there exists a state $s'$ with $T_b(s,s')$, such that either $s'\in\win_b$, 
or for all states $s''$ with $T_w(s',s'')$, $s''\not\in\win_w$ and $s''\in \WIN_k$.
\end{itemize}
We say that Black has a winning strategy of depth $d$, if the initial state $s_0\in \WIN_d$.
\end{definition}


\section{QBF encoding}
\label{sec:qbfencoding}

\subsection{Prefix}
\label{sybsec:prefix}

\begin{table}[b]
\centering
\begin{tabular}{l|l}
\hline
Variable & Description \\
\hline
$\action^i$ & $\ceil{\log(|\action_b|)}$ / $\ceil{\log(|\action_w|)}$ vars for B's/W's action if $i$ is odd/even\\
$\gamestop^i$ & game-stop variable at time step $i$\\
$\X^{i},\Y^{i}$ & $\ceil{\log(m+1)},\ceil{\log(n+1)}$ variables for the action parameters\\
$\legalbound^i$ & white legal bound indicator variable at time step $i$\\
$\preflags^i$ & max \#W-preconditions indicator variables for White preconditions at time $i$\\
$\B_x,\B_y$/$\W_x,\W_y$ & $\ceil{\log(m+1)},\ceil{\log(n+1)}$ variables for Black's/White's goal position\\
$\B_c$/$\W_c$ & $\ceil{\log(|\goal_b|)}$/$\ceil{\log(|\goal_w|)}$ variables for Black's/White's goal index\\
$\W_{ce}$ & $\ceil{\log(\mbox{max \#W-goal subconditions})}$, counter-example to White's goal $\W_c$\\
$\sympos_x,\sympos_y$ & $\ceil{\log(m+1)},\ceil{\log(n+1)}$ variables indicating a symbolic board position\\
$o^i, w^i$ & two variables indicating the state at the symbolic board position at time $i$\\
\hline
\end{tabular}
\medskip
\caption{Encoding variables and descriptions}
\label{table:vars}
\end{table}

\noindent
We now present a uniform encoding in QBF of ``there exists a winning strategy for Black of at most $d$ moves'',
given a fixed game domain and problem instance in BDDL.
Following the descriptions of variables from Table \ref{table:vars},
we first introduce variables for black and white moves.
Black's moves are existential whereas White's moves are universal.
We represent a move by using action $\action$ and index variables $\X,\Y$ at each time step.
To handle games shorter than $d$ steps, we introduce game-stop variables ($\gamestop$).

Since a white move is encoded with universal variables, some encoded moves may not satisfy the required constraints.
To remember if a played white move is valid, we use existential indicator variables $\legalbound$ and $P$.
Here $P$ will mean that the preconditions for White's move $A$ on $(X,Y)$ hold and
$\legalbound$ will mean that $A$, $X$ and $Y$ stay within legal bounds.
Note that, in each specific universal branch of a white move the indicator variables are completely determined
by the chosen move and the state of the board.
%
\begin{align*}
\underbrace{\exists \action^1, \X^1, \Y^1, \gamestop^1}_{\mbox{Black move 1}} \quad
\underbrace{\forall \action^{2}, \X^{2}, \Y^{2},\gamestop^2}_{\mbox{White move 2}} \quad
\underbrace{\exists\legalbound^{2}, \preflags^{2}}_{\mbox{White Indicators 2}} \dots \quad
\underbrace{\exists \action^{d}, \X^{d}, \Y^{d}, \gamestop^{d}}_{\mbox{Black move d}}
\end{align*}

Next, we introduce the variables to check for black and white goals.
For Black, we need to show that there exists a position $\B_x, \B_y$ on the board that satisfies one
of Black's goal conditions, $\goal_b[\B_c]$.
For White, we need to show absence of the white goal condition, i.e., 
for every board position $\W_x, \W_y$ and for each white goal condition $\goal_w[\W_c]$,
we need to show there exists a sub-condition $\goal_w[\W_c][\W_{ce}]$ that is violated (i.e., a counter example).

Finally, we introduce variables to check the validity of all the moves.
Universal variables $\sympos_x, \sympos_y$ represent a symbolic position on the board.
These are used to check the preconditions and effects on each board position in a uniform manner.
At each time step, the state variables $\open, \white$ represent if the symbolic position $(\sympos_x, \sympos_y)$ is open and/or
white, respectively; so black corresponds to $\open\wedge \neg \white$.
The symbolic and state variables together represent a full state of the board at any time step.
%
\begin{align*}
\underbrace{\exists \B_x, \B_y, \B_c}_{\mbox{Black goal}} \quad 
\underbrace{\forall \W_x, \W_y, \W_c \exists \W_{ce}}_{\mbox{White goal}} \quad
\underbrace{\forall \sympos_x, \sympos_y}_{\mbox{symbolic pos}} \quad
\underbrace{\exists \open^1, \white^1, \dots, \open^{d+1}, \white^{d+1}}_{\mbox{state variables}}
\end{align*}

\subsection{Matrix}
\label{subsec:matrix}

The matrix of our QBF encoding consists of an initial constraint and a constraint on moves.
\begin{align*}
\Icon \land B^{1}
\end{align*}
$\Icon$ (initial state), $\move_b^{i}$, $\move_w^{i}$ (valid moves),
$\Gcon_b^{i}$, and $\Gcon_w^{i}$ (goal checks) will be introduced
in subsequent sections.
We now recursively define the turn based player constraints as follows:
\begin{align*}
&B^{i} := \move_b^{i} \land (\gamestop^{i} \implies \Gcon_b^{i+1}) \land (\neg\gamestop^{i} \implies W^{i+1}) \mbox{, for $i=1, 3, \ldots, d-2$}\\
&W^{i} := \move_w^{i} \land 
((\legalbound^{i} \land \bigwedge_{p \in \preflags^{i}} p)
 \implies 
((\gamestop^{i} \implies \Gcon_w^{i+1}) \land
(\neg\gamestop^{i} \implies B^{i+1})))  \mbox{, for $i=2, 4, \ldots, d-1$}\\
&B^{d} := \move_b^{d} \land \Gcon_b^{d+1}
\end{align*}

\begin{itemize}
\item In Black's turn (odd $i$), the black move must be valid ($\move^i_b$). If the game is stopped ($\gamestop$),
the black goal is enforced ($\Gcon^{i+1}_b$), else we continue with White's turn $W^{i+1}$.
\item In White's turn (even $i$), the white move constraints must hold ($\move^i_w$). 
$\move^i_w$ also specifies
the indicator variables $\legalbound^i$ and $\preflags^i$. Note that we only care about valid moves
(if White can only play invalid, Black has won). 
If the game is stopped, the white goal is enforced  ($\Gcon^{i+1}_w$), else we continue with Black's turn ($B^{i+1}$).
\item At time step $d$, a valid black move should lead to the black goal condition in time $d+1$.
\end{itemize}


%

\subsection{Defining Initial, Move and Goal Circuits}
\label{par:auxiliarycircuits}

\paragraph*{Auxiliary Circuits}
\label{subsec:auxiliarycircuits}

In our QBF encoding, we generate three auxiliary constraints:
\begin{itemize}
\item Compute relative symbolic index for a sub-condition ($\symbimp$).
\item Compute state constraints for a sub-condition ($\subcon$).
\item Compute bound constraints for a condition ($\boundcircuit$).
\end{itemize}
The key idea of the relative symbolic index is to test if the parameters $(?x,?y)$ that
occur in some sub-condition $p(e1,e2)$
refer to the current value of the symbolic position variables $(S_x,S_y)$. If so, the 
current state $(o,w)$ will be forced to the value indicated by $p$.
For the conditions, also bound constraints are generated to avoid out-of-bound references.

\medskip
Before we can define these constraints more precisely, we first need to shortly explain how we encoded
some arithmetic sub-circuits, in particular $\adder$, $\subtractor$, $\compute$,
$\lessthan$ and $\equality$.

In sub-conditions, we allow addition and subtraction with a constant value.
Given a sequence of Boolean variables $\vars$ and an integer $k$, we generate an adder circuit $\adder(\vars,k)$ with $\vars$ as input gates and a sequence of output gates 
for the binary representation of $\vars+k$.
For subtracting, we apply an adder circuit using 2's complement i.e., $\subtractor(\vars,k) := \adder(\vars,k')$, where $k'$ is a 2's-complement of $k$.
Given an expression $E = (e1,e2)$, we use an \emph{Adder-Subtractor} generator $\compute(\vars,e1)$ to generate index constraints on $?x$ and $\compute(\vars,e2)$ to generate index constraints on $?y$.
The generator $\compute(\vars,e)$ returns the sub-circuit:
\begin{itemize}
\item $\adder(\vars,k)$ if $e$ is ?x+$k$ or ?y+$k$; $\subtractor(\vars,k)$ if $e$ is ?x-$k$ or ?y-$k$;
\item $V$, if $e$ is $?x$ or $?y$; and $e$, if $e$ is an integer.
\end{itemize}

The equality generator $\equality(p,p')$ generates (1) an equality circuit with $p$ and $p'$ as inputs if both are sequence of variables; (2) a single And gate if either $p$ or $p'$ is an integer.
In both cases there is only one output gate which is true iff the inputs are equal.

The less-than comparator circuit $\lessthan(\vars,k)$ for integer $k$ takes as input a sequence of 
Boolean variables $\vars$ and has a single output gate which is true iff the binary input is less than $k$. Lower bounds (greater-than-or-equal) can be achieved using the negation of less-than.

\medskip
We can now define the circuits for the relative index, the sub-condition constraint, and the bound constraints.
\begin{definition}
Given a sub-condition $p(e1,e2)$, sequences of variables $\vars_x,\vars_y$ representing a position on the board.
We define a constraint generator, \emph{Relative Index}
\begin{align*}
&\symbimp(\vars_x,\vars_y,p(e1,e2)) := \equality(\compute(\vars_x,e1),\sympos_x) \land \equality(\compute(\vars_y,e2),\sympos_y)
\end{align*}
\end{definition}

\begin{definition}
Given a sub-condition $p(e_1,e_2)$ and state variables ${\open,\white}$, we define sub-condition constraint
\begin{align*}
\subcon(p(e_1,e_2),\open,\white) :=
\neg\open \land \neg\white \mbox{, if $p=\blacksymbol$}; \quad
\neg\open \land \white \mbox{, if $p=\whitesymbol$};  \quad
\open \mbox{, if $p=\opensymbol$} 
\end{align*}
We negate the sub-condition constraint if the sub-condition $p(e1,e2)$ is negated.
\end{definition}

\begin{definition}
Given a condition $C$, and sequences of variables $\vars_x, \vars_y$ representing a position. 
Assume $\bounds(C) = [l_x,u_x] \times [l_y, u_y]$.
We define a bound generator:
\begin{align*}
    \boundcircuit(\vars_x,\vars_y,C) := \;
& \neg\lessthan(\vars_x,l_x) \land \lessthan(\vars_x,u_x+1)\land \neg\lessthan(\vars_y,l_y) \land \lessthan(\vars_y,u_y+1)
\end{align*}
\end{definition}
\noindent The output gate of circuit is true all upper and lower bounds hold.

For example,
consider a sub-condition $\opensymbol(?x,?y+1)$ in precondition from Evader-Pursuer (see Listing \ref{lst:evaderpursueraction}).
In our encoding at time $1$ we generate a corresponding constraint:
\begin{align*}
&\symbimp(\X^1,\Y^1,\opensymbol(?x,?y+1)) \implies\subcon(\opensymbol(?x,?y+1),\open^1,\white^1)
\end{align*}
This is equivalent to generating the circuit
\begin{align*}
\equality(\X^1,\sympos_x) \land \equality(\adder(\Y^1,1),\sympos_y) \implies \open^1
\end{align*}
This circuit is true if the open predicate in the relative symbolic branch $(?x,?y+1)$ is true.
To give a concrete example, if the first black move is played on position $(\X^1,\Y^1)=(1,1)$ then in the symbolic branch $(\sympos_x,\sympos_y)=(1,2)$ the open predicate is implied.

For the same action, we enforce the bound constraints using $\boundcircuit(\X^1,\Y^1,\all(\text{down-two}))$ which is equivalent to generating the circuit
\begin{align*}
&\neg\lessthan(\X^1,\xmin) \land \lessthan(\X^1,\xmax+1) \land \neg\lessthan(\Y^1,\ymin) \land \lessthan(\Y^1,\ymax-1)
\end{align*}
Every position where $y$ is less than $\ymax-1$ is a legal position for down-two action.

\paragraph*{Initial State $\Icon$} Given the initial specification $\I$ (as in Def.~\ref{def:gameinstance}), 
we specify that the state $(o^1,w^1)$ at time step 1 has the proper value for symbolic position
$(\sympos_x,\sympos_y)$.
\begin{align*}
\Icon := \;
&((\bigvee_{\blacksymbol(e1,e2)\in\I} \equality(e1,\sympos_x) \land \equality(e2,\sympos_y)) \implies \neg\open^1 \land \neg\white^1)\nonumber\\
&\land((\bigvee_{\whitesymbol(e1,e2)\in\I} \equality(e1,\sympos_x) \land \equality(e2,\sympos_y)) \implies \neg\open^1 \land \white^1)\nonumber\\
&\land(\bigwedge_{p(e1,e2)\notin\I} \equality(e1,\sympos_x) \land \equality(e2,\sympos_y) \implies \open^1)
\end{align*}

\paragraph*{Black Move $\move_b^{i}$}
The following constraints specify that the black action $A^i(X^i,Y^i)$ at time step $i$ is a legal move,
with respect to the state transition $(\open^i,\white^i)\to (\open^{i+1},\white^{i+1})$ of the current symbolic $(\sympos_x,\sympos_y)$-branch.
\begin{align*}
\move_b^{i} := \;
&\lessthan(\action^i,\mid\action_b\mid) \land \bigwedge_{j=1}^{\mid\action_b\mid} \big(\equality(\action^i,j) \implies & \text{ index bound on conditions} \nonumber\\
&\quad\big(\boundcircuit(\X^i,\Y^i,\all(\action_b[j]))\quad\land &\text{ index bound subconditions}\nonumber\\
&\quad\bigwedge_{C\in \pre(\action_b[j])} \symbimp(\X^i,\Y^i,C)\implies\subcon(C,\open^i,\white^i)\quad\land &\text{ preconditions hold at } i\nonumber\\
&\quad\bigwedge_{C\in \eff(\action_b[j])} \symbimp(\X^i,\Y^i,C)\implies\subcon(C,\open^{i+1},\white^{i+1})\quad\land &\text{ effects hold at } i+1\nonumber\\
&\quad\big(\neg\bigvee_{C\in \eff(\action_b[j]) } \symbimp(\X^i,\Y^i,C)\big)\implies&\text{unchanged positions}\nonumber\\
&\quad\qquad(\equality(\open^{i},\open^{i+1}) \land \equality(\white^{i},\white^{i+1})\big)\big)&\text{are propagated}
\end{align*}
\paragraph*{White Move $\move_w^{i}$}
For white moves at time step $i$, we first specify that 
the legal bound variable is true iff for each action the bound constraints hold.
For each action, we also set the precondition flags $\preflags^{i}$ to true iff the preconditions hold\footnote{For simplicity of the presentation, we assume that all white actions have the same number of preconditions.}.
Since the preconditions refer to positions in different symbolic branches, we set the flags after each move so that the values are available in all symbolic branches.
The effects should hold at time step $i+1$ for each action, if the bounds are legal and preconditions are true.
The positions that are not changed by the legal move are propagated.
\begin{align*}
\move_w^{i} := \;
&\big(\big(\bigwedge_{j=1}^{\mid\action_w\mid} \equality(\action^i,j) \implies \boundcircuit(\X^i,\Y^i,\all(\action_w[j]))\land\lessthan(\action^i,\mid\action_w\mid)\big) \iff \legalbound^i\big)\quad\land\nonumber\\
&\big(\bigwedge_{j=1}^{\mid\action_w\mid} \equality(\action^i,j) \implies\nonumber\\
&\quad\bigwedge_{k=1}^{\mid\preflags\mid} \symbimp(\X^i,\Y^i,\pre(\action_w[j])[k])\implies(\subcon(\pre(\action_w[j])[k],\open^{i},\white^{i}) \iff \preflags^{i}[k])\quad\land\nonumber\\
&\quad\big(\legalbound^{i} \land \bigwedge_{p \in \preflags^{i}} p \implies\bigwedge_{C\in \eff(\action_w[j])} \symbimp(\X^i,\Y^i,C)\implies\subcon(C,\open^{i+1},\white^{i+1})\quad\land\nonumber\\
&\quad\qquad\big(\neg\bigvee_{C\in \eff(\action_w[j])} \symbimp(\X^i,\Y^i,C)\big)\implies(\open^{i}\iff\open^{i+1}) \land (\white^{i} \iff \white^{i+1})\big)\big)
\end{align*}
\paragraph*{Black Goal $\Gcon_b^{i}$}
We check if Black meets its $\B_c$'s goal condition at position $(\B_{x},\B_{y})$ at time $i$. 
Since goal conditions are encoded in binary, we need to restrict $\B_c$ within boundaries.
To check the goal at time step $i$, we first propagate the state variables to last time step $d+1$.
Next, we check the goal condition at $d+1$. This avoids copies of black goal per time step.
\begin{align*}
\Gcon_b^{i} := \; &
\lessthan(\B_c,\mid\goal_b\mid) \land
\equality(\open^{i},\open^{d+1}) \land \equality(\white^{i},\white^{d+1})  \quad\land\nonumber\\
&\bigwedge_{j=1}^{\mid\goal_b\mid} \big( \equality(\B_c,j) \implies \boundcircuit(\B_{x},\B_{y},\goal_{b}[j]) \quad\land{}\nonumber\\
&\qquad\bigwedge_{C\in \goal_{b}[j]} \big(\symbimp(\B_{x},\B_{y},C) \implies\subcon(C,\open^{d+1},\white^{d+1})\big)\big)
\end{align*}
\paragraph*{White Goal $\Gcon_w^{i}$}
For White's goal, we check that $\W_{ce}$ is a counter-example to White's goal condition $\W_c$
at position $(\W_{x},\W_{y})$ at time step $i$.
As for the black goal, we first propagate the state variables to the last time step $d+1$.
White's goal is violated if $(\W_{x},\W_{y})$ are out-of-bounds, or if $\W_{ce}$ is a legal
index and the corresponding sub-constraint in $\W_c$ is violated.
\begin{align*}
\Gcon_w^{i} :=  \; &
\equality(\open^{i},\open^{d+1}) \land \equality(\white^{i},\white^{d+1}) \quad\land \nonumber\\
&\bigwedge_{j=1}^{\mid\goal_w\mid} \big(\equality(\W_c,j) \land \boundcircuit(\W_{x},\W_{y},\goal_{w}[j]) \implies \big(\lessthan(\W_{ce},\mid\goal_w[j]\mid)\quad \land \nonumber\\
&\quad\bigwedge_{k=1}^{\mid\goal_w[j]\mid} \big(\symbimp(\W_{x},\W_{y},\goal_w[j][k]) \land  \equality(\W_{ce},k)\implies\neg\subcon(\goal_w[j][k],\open^{d+1},\white^{d+1})\big)\big)\big)
\end{align*}

\section{Implementation and Evaluation}
\label{sec:implementation}

We provide an open source implementation for the BDDL to QBF translation (in QCIR format, which
can be translated to QDIMACS). All the BDDL models for games described, benchmarks and data are available online%
\footnote{https://github.com/irfansha/Q-sage}. One can easily model other games in BDDL and generate
QBF formulas to solve those games up to a depth and extract a winning strategy.

For experimental evaluation, we consider small boards for various games described in this paper.
We generate QBF instances and try to solve them with
the QDIMACS solvers DepQBF \cite{DBLP:conf/cade/LonsingE17} and CAQE \cite{DBLP:conf/fmcad/RabeT15} and with the QCIR solvers Quabs (QU)
\cite{DBLP:journals/corr/Tentrup16} and CQESTO (CT) \cite{DBLP:conf/sat/Janota18}.
We also applied QBF preprocessors Bloqqer (B) \cite{bloqqer} and HQSpre (H) \cite{DBLP:journals/jsat/WimmerSB19} on the QDIMACS encodings. 
We give a 1 hour time limit and 8GB memory limit (preprocessing+solving) for each instance. All computations for the experiments are run on a cluster.%
\footnote{\url{http://www.cscaa.dk/grendel-s}, each problem uses one core on a
Huawei FusionServer Pro V1288H V5 server, with 384 GB main memory and
48 cores of 3.0 GHz (Intel Xeon Gold 6248R).}

In Table~\ref{table:allresults}, we report the critical depths in \textbf{bold} when a winning strategy is found by any solver-preprocessor combination.
We report the non-existence of winning strategies in plain font if the depth-bound is known to be complete.
Otherwise, we show the maximum refuted depth for the unsolved instances in \emph{italic}.

\paragraph*{Positional games} In positional games for HTTT, we consider benchmarks from \cite{DBLP:conf/sat/DiptaramaYS16}.
In Table \ref{table:allresults}(a), we solve all shapes (standard names) on a 3x3 board. On a 4x4 board,
we solve all shapes except skinny and knobby.
For Hex games, we use Hein's benchmarks as in \cite{10.1007/978-3-030-51825-7_31}.
For these, we first generate simplified BDDL problem files with explicit winning sets. We drop explicit winning sets with more than the number of black moves.
In Table \ref{table:solverdata}, we report the number of instances solved. Our simplified instances perform on par with the COR encoding by \cite{10.1007/978-3-030-51825-7_31}.

\begin{table}[tb]
\centering
\begin{tabular}{l | llllllll}
\multicolumn{9}{c}{(a) Harary's Tic-Tac-Toe (HTTT)}\\
\hline
& \multicolumn{8}{|c}{Several standard shapes}\\
\hline
n  &  D & E & Ey & F & K & S & T & Tp\\
\hline
3 & \textbf{3} & \textbf{5} & 9 & 9 & 9 & - & 9 & \textbf{9}\\
4 & \textbf{3} & \textbf{5} & \textbf{7} & 15 & \textit{11} & \textit{13} & \textbf{5} & \textbf{9}\\
\hline
\end{tabular}
~~
\begin{tabular}{llll|lll}
\multicolumn{7}{c}{(b) Breakthrough, for first and second player}\\
\hline
& \multicolumn{3}{c|}{first player} & \multicolumn{3}{c}{second player}\\
\hline
m / n &  4 & 5 & 6 & 4 & 5 & 6\\
\hline
2 & 13 & \textit{17} & \textbf{15} & \textbf{8} & \textbf{10} & \textit{14}\\
3 & 19 &\textit{11} & \textit{9}& \textbf{12} &\textit{12} & \textit{10}\\
\hline
\end{tabular}
\medskip

\begin{tabular}{llll}
\multicolumn{4}{c}{(c) Connect-c}\\
\hline
nxn &  C-2 & C-3 & C-4\\
\hline
2x2 & \textbf{3} & - & -\\
3x3 & \textbf{3} & 9 & -\\
4x4 & \textbf{3} & \textbf{9} & 15\\
5x5 & \textbf{3} & \textbf{9} & \textit{11}\\
6x6 & \textbf{3} & \textbf{9} & \textit{11}\\
\hline
\end{tabular}
~~
\begin{tabular}{lllllll}
\multicolumn{6}{c}{(d) Domineering}\\
\hline
n / m &  2 & 3 & 4 & 5 & 6\\
\hline
2 & \textbf{2} & \textbf{2} & 5 & \textbf{6} & \textbf{6}\\
3 & \textbf{4} & \textbf{4} & 7 & 8 & 10\\
4 & \textbf{4} & \textbf{6} & \textbf{8} & \textbf{10} & \textbf{12}\\
5 & 6 & \textbf{8} & 11 & 13 & \textit{11}\\
6 & \textbf{6} & \textbf{6} & \textbf{12} & \textit{11} & \textit{11}\\
\hline
\end{tabular}
~~
\begin{tabular}{llll}
\multicolumn{4}{c}{(e) Evader-Pursuer (dual)}\\
\hline
nxn &  P & EP & EP-d\\
\hline
4x4 & (1,2) & $\textit{21}^{*}$ & \textbf{2}\\
4x4 & (2,3) & \textbf{3} & \textit{10}\\
8x8 & (2,3) & \textit{11} & \textbf{6}\\
8x8 & (3,4) & \textbf{7} & \textit{10}\\
\hline
\multicolumn{4}{l}{${}^*$We solved 21, but not 19}\\
\end{tabular}
\medskip
\caption{Results for HTTT, Breakthrough, Connect-c, Domineering and Evader-Pursuer}
\label{table:allresults}
\end{table}

\paragraph*{Breakthrough} Boards up to 6x5 were solved in \cite{DBLP:conf/acg/SaffidineJC11} using Proof Number Search and handcrafted race patterns.
As far as we know, we provide the first encoding of BreakThrough in QBF.
On an $m{\times}n$ board, we need $4mn-10m+1$ moves to show that the first player has no winning strategy.
On the other hand, a second player winning strategy, i.e., giving up the first move, may be demonstrated much earlier.
In our experiments, Table \ref{table:allresults}(b), we search for both first and second player winning strategies.
For the first player, we prove non-existence for boards 2x4 and 3x4 and existence for 2x6.
For the second player, we show existence for boards 2x4, 2x5 and 3x4 (all optimal strategies).
Partial endgame tablebases were used \cite{10.1007/978-3-319-50935-8_1} for improving a 6x6 Breakthrough playing program. One could also use QBF for such shallow endgames.

\paragraph*{Connect-c} The first QBF encoding for Connect-c \cite{Gent2003EncodingCU} 
restricted the search to columns.
The authors reported that no winning strategies exist for Connect-4 on a 4x4 board.
In Table \ref{table:allresults}(c), we consider $c\in{2,3,4}$ and nxn board sizes up to n=6.
Connect-2 and Connect-3 have small critical depths and can be solved easily even for large board sizes. For Connect-4, we can solve the 4x4 board within 1 hour.
By relaxing action definitions in BDDL, one could improve the model to allow only column search.
This would probably solve larger boards.

\paragraph*{Evader and Pursuer} We consider the instances from \cite{Ansotegui2005AHQ}. In an nxn grid, the Evader starts at (xmax,ymin) and the goal is to reach (xmin,ymin).
Pursuer (column $P$) starts at different places. \cite{Alur2004SymbolicCT} designed the instances for early termination, so all are UNSAT.
We also add SAT instances with different Pursuer positions.
In the first instance in Table \ref{table:allresults}(e), Evader wins after one white move. Ideally, QBF solvers should be able to infer this quickly.
For depths ${17,21}$, DepQBF with Bloqqer indeed finds unsatisfiability within seconds. Surprisingly, 
for depths $15, 19$ it ends up searching unnecessary search space.
Among other solvers, only CQESTO scales up to depth $17$ (solves within seconds). However, it times out on deeper games.
We consider a dual version of Evader-Pursuer as well: instead of search for Evader winning, we search for Pursuer winning. Here, Pursuer wins if it reaches the goal first or kills the Evader.

\paragraph*{Domineering} In table \ref{table:allresults}(d), we report results on board sizes up to 6x6.
Domineering on rectangular boards has been solved mainly by combinatorial theory non-optimally \cite{Lachmann2000Domineering},
perfect solving i.e., without any search \cite{DBLP:conf/ijcai/Uiterwijk13} and domain specific solvers \cite{DBLP:conf/cg/Uiterwijk16}.
11x11 has been solved to show first player winning (took 174 days and 15 h on a standard desktop computer) by \cite{DBLP:conf/cg/Uiterwijk16}.
As far as we know, we provide the first encoding of Domineering in QBF.
Our results are consistent with the results in literature; we also provide critical depths, i.e., optimal winning strategies.

\paragraph*{Comparing Multiple QBF Solvers and Preprocessors}
In Table \ref{table:solverdata}, we report results from different QBF solvers on all solved instances.
Overall, DepQBF with Bloqqer performs well and solves most instances.
If we look at the clauses generated by Bloqqer, it adds long clauses with indicator variables for early backtracking in case of an illegal move.
On the other hand, HQSpre does not seem to fully recover this information with shorter clauses.
Interestingly, there are some unique instances solved with HQSpre.
The performance of CAQE is interesting: it performs well for the domains like positional games where there is a low branching factor for moves.
On the contrary, for the domains like Breakthrough and Evader-Pursuer, Circuit solvers outperform CAQE.
The circuit solver CQESTO uniquely solves an 8x8 instance in the Evader-Pursuer domain. On the other hand, it performs badly on the Hex domain.
We suspect that this might be due to lack of early pruning in Hex, where the solver often needs to explore until the innermost quantifier to conclude satisfiability or unsatisfiability.
In the Domineering domain, all solvers perform relatively well, perhaps due to very simple action and goal configurations.
Since different domains have a different number of instances, we also report the average coverage per domain, In this weighted measure, circuit solvers perform on par with CAQE.
These results suggest that the domains encoded using BDDL provide a rich set of benchmarks.

\begin{table}[tb]
\centering
\begin{tabular}{l|ll|ll|l|l}
\hline
& \multicolumn{2}{c|}{DepQBF} & \multicolumn{2}{c|}{CAQE} & CT & QU\\
Dom / SP & B & H & B & H & & \\
\hline
HTTT & \textbf{15} & 12 & 12 & 11 & 10 & 11 \\
Hex & \textbf{26} & 18 & 22 & 19 & 10 & 18\\
B & \textbf{6} & 4 & 2 & 2 & 2 & 3\\
B-SP & \textbf{6} & 5 & 2 & 3 & 4 & 4\\
C-c & \textbf{12} & \textbf{12} & \textbf{12} & 8 & 8 & 8\\
EP & \textbf{3} & 2 & 2 & 2 & \textbf{3} & 2\\
EP-dual & 1 & \textbf{3} & 1 & 2 & 2 & \textbf{3}\\
D & \textbf{25} & 20 & 23 & 19 & 22 & 23\\
\hline
total & \textbf{94} & 76 & 76 & 66 & 61 & 72\\
Avg. Cov. & \textbf{0.86} & 0.74 & 0.61 & 0.58 & 0.6 & 0.67\\
\hline
\end{tabular}
\caption{Number of instances solved by various solver-preprocessor combinations. Average Coverage
indicates the fraction of solved instances by a combination, averaged per game row.}
\label{table:solverdata}
\end{table}

\subsection{Winning Strategy Validation}
\label{subsec:correctnessandvalidation}
In case of classical planning, one can use an external plan validator for correctness of plans.
Validation of winning strategies from QBF is more complex, errors can occur in modelling, encoding or solving.
Testing if a winning strategy exists for already solved game instances and testing if the winning move is correct can be used to some extent.
However, this is not sufficient: The encoding can only be validated properly by looking at the values of variables.

\begin{figure}[tb]
\centering
\scalebox{.8}{%
\begin{tikzpicture} [node distance=1.2cm]
\node (in1) [io] {QBF Cert.};
\node (pro1) [process, below of=in1] {SAT solver};
\node (out1) [io, below of=pro1] {Assignment \& B. move};
\node (isgoal) [decision, right of=out1, xshift=2cm] {Goal};
\node (stop1) [startstop, below of=isgoal,yshift=-0.5cm] {Valid};
\node (isend) [decision, right of=isgoal, xshift=1.2cm] {k=0};
\node (stop2) [startstop, below of=isend, yshift=-0.5cm] {Invalid};
\node (pro2) [process, above of=isend, yshift=1cm] {Interactive tool};
\node (out2) [io, left of=pro2,xshift=-1cm, yshift=-1cm] {Assumptions \& W. move};
\draw [arrow] (in1) --node[anchor=west] {k=d} (pro1);
\draw [arrow] (pro1) --node[anchor=west] {k=k-1} (out1);
\draw [arrow] (out1) -- (isgoal);
\draw [arrow] (isgoal) --node[anchor=west] {yes} (stop1);
\draw [arrow] (isgoal) --node[anchor=north] {no} (isend);
\draw [arrow] (isend) -- node[anchor=west] {yes} (stop2);
\draw [arrow] (isend) -- node[anchor=west] {no} (pro2);
\draw [arrow] (pro2) -|node[anchor=south] {k=k-1} (out2);
\draw [arrow] (out2) -- (pro1);
\end{tikzpicture}
}
\caption{Winning strategy validation with Interactive play}
\label{fig:validation}
\end{figure}
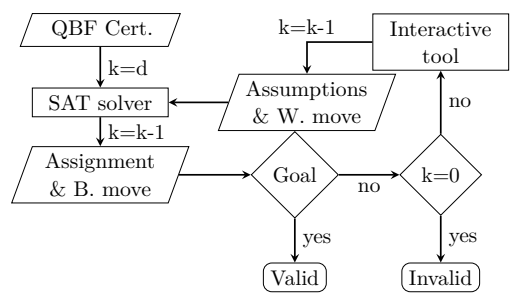

We propose to use QBF certificates, as generated by some QBF solvers.
By giving assumptions for universal variables to a SAT solver, along with the certificate,
one can extract the values of existential variables, i.e., black moves and board state variables.
We interactively play as a white player with the winning strategy as an opponent, and visualize the board at each time step.
At the end of the play, we check if the output goal gate is true, i.e., Black won the game.
Illegal white moves are identified by checking the outputs of certain intermediate gates.
Fig.~\ref{fig:validation} shows the flow chart for our QBF-based validation framework.
Appendix \ref{sec:valdemo} presents a demo of validating a strategy for the Tic game. 
We visualize the board by extracting values for state variables from the certificate.

\section{Conclusion and Future Work}
\label{sec:conclusionandfuturework}

We propose a compact modelling format BDDL for a subset of 2-player games played on a grid.
Several classical games are modelled, such as positional games, Connect4, BreakThrough, Evader and Pursuer, and Domineering.
We provide a concise QBF translation from BDDL to QBF for generating winning strategies, and provide an open source implementation for the translation.
This provides the first QBF encoding for the games BreakThrough, KnightThrough and Domineering.
Using existing QBF solvers and preprocessors, we solved several small instances of these games,
yielding concrete winning strategies.
This provides a rich set of benchmarks for QBF solvers. Hopefully, such diverse QBF instances can help the QBF solving community.
We also provided a framework for validating our winning strategies by using QBF certificates.

As of now, our action and goal conditions are somewhat restricted.
One research direction would be to relax these restrictions. For example, allowing static predicates can extend our models with multiple pieces.
In fact, with static predicates one could already model several chess puzzles. Allowing non-static predicates can result in better domains for games like Connect4.
Relaxing goal conditions can help to model more complex games implicitly.
Our validation of QBF encodings is especially useful when checking alternative game encodings.
QBF solvers that generate certificates are several orders slower than their counterparts and can blow up in size even for small depths.
There is a need to improve certificate generation for extensive validation.



\bibliography{../../references}

\appendix

\section{A demo for winning strategy validation}
\label{sec:valdemo}
\begin{figure}[h]
\centering
\includegraphics[scale=0.2]{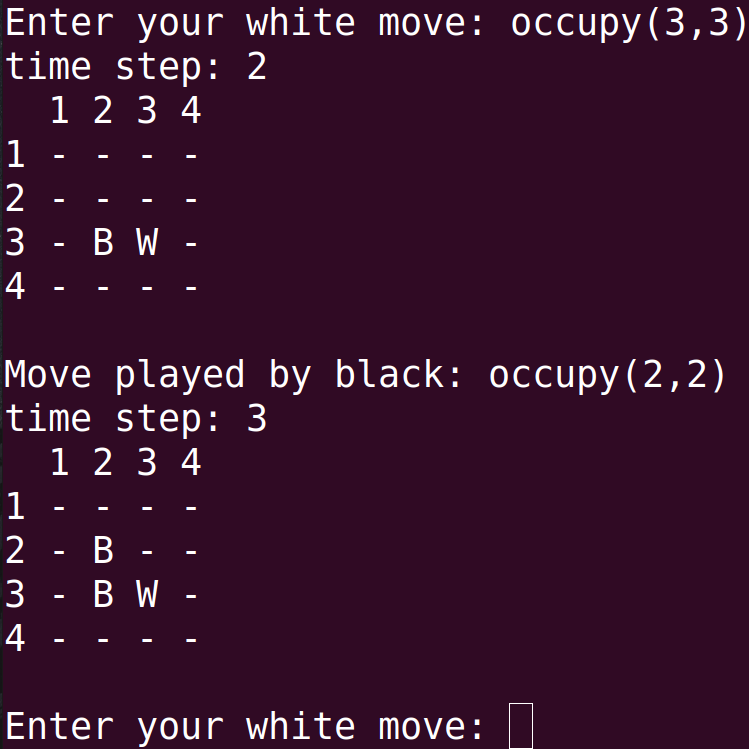}
\caption{A snapshot of winning strategy validation}
\label{fig:val}
\end{figure}

We provide a script for winning strategy validation demo for HTTT Tic problem.
One could play with the winning strategy to validate various game plays.
Use the following command for demo in our tool Q-sage:
\begin{verbatim}
python3 general_interactive_play.py
\end{verbatim}
In Figure \ref{fig:val}, we provide a screenshot of interactive play during validation.

\section{Complete Models for Classical Games}
\label{sec:completemodels}
In this Section, we provide complete models for classical games described in the main paper.
\begin{itemize}
\item Listings \ref{lst:connect4actions},\ref{lst:connect4problem} model an example Connect4 game.
\item Listings \ref{lst:breakthroughdomain},\ref{lst:breakthroughproblem} model an example BreakThrough game.
\item Listings \ref{lst:ktdomainsnippets},\ref{lst:ktproblem} model an example KnightThrough game. We have omitted 12 of 16 actions in the domain, the full domain is available in our tool.
\item Listings \ref{lst:evdomainsnippets},\ref{lst:evproblem} model an example Evader-Pursuer game. We have omitted 14 of 18 actions in the domain, the full domain is available in our tool.
\item Listings \ref{lst:domineeringdomain},\ref{lst:domineeringproblem} model an example Domineering game.
\end{itemize}

\begin{lstlisting}[caption={Connect4: Domain}, label={lst:connect4actions},language=stripspddl,mathescape]
#blackactions
:action occupyOnTop
:parameters (?x,?y)
:precondition (open(?x,?y) NOT(open(?x,?y+1)))
:effect (black(?x,?y))
:action occupyBottom
:parameters (?x,?y)
:precondition (open(?x,ymax))
:effect (black(?x,ymax))
#whiteactions
:action occupyOnTop
:parameters (?x,?y)
:precondition (open(?x,?y) NOT(open(?x,?y+1)))
:effect (white(?x,?y))
:action occupyBottom
:parameters (?x,?y)
:precondition (open(?x,ymax))
:effect (white(?x,ymax))
\end{lstlisting}

\begin{lstlisting}[caption={Connect4: Example Problem}, label={lst:connect4problem},language=stripspddl,mathescape]
#boardsize
7 6
#init
#depth
9
#blackgoal
(black(?x,?y) black(?x+1,?y) black(?x+2,?y) black(?x+3,?y))
(black(?x,?y) black(?x,?y+1) black(?x,?y+2) black(?x,?y+3))
(black(?x,?y) black(?x+1,?y+1) black(?x+2,?y+2) black(?x+3,?y+3))
(black(?x,?y) black(?x+1,?y-1) black(?x+2,?y-2) black(?x+3,?y-3))
#whitegoal
(white(?x,?y) white(?x+1,?y) white(?x+2,?y) white(?x+3,?y))
(white(?x,?y) white(?x,?y+1) white(?x,?y+2) white(?x,?y+3))
(white(?x,?y) white(?x+1,?y+1) white(?x+2,?y+2) white(?x+3,?y+3))
(white(?x,?y) white(?x+1,?y-1) white(?x+2,?y-2) white(?x+3,?y-3))
\end{lstlisting}

\begin{lstlisting}[caption={Breakthrough: Domain}, label={lst:breakthroughdomain},language=stripspddl,mathescape]
#blackactions
:action forward
:parameters (?x,?y)
:precondition (black(?x,?y) open(?x,?y-1))
:effect (open(?x,?y) black(?x,?y-1))
:action left-diagonal
:parameters (?x,?y)
:precondition (black(?x,?y) NOT(black(?x-1,?y-1)))
:effect (open(?x,?y) black(?x-1,?y-1))
:action right-diagonal
:parameters (?x,?y)
:precondition (black(?x,?y) NOT(black(?x+1,?y-1)))
:effect (open(?x,?y) black(?x+1,?y-1))
#whiteactions
:action forward
:parameters (?x,?y)
:precondition (white(?x,?y) open(?x,?y+1))
:effect (open(?x,?y) white(?x,?y+1))
:action left-diagonal
:parameters (?x,?y)
:precondition (white(?x,?y) NOT(white(?x-1,?y+1)))
:effect (open(?x,?y) white(?x-1,?y+1))
:action right-diagonal
:parameters (?x,?y)
:precondition (white(?x,?y) NOT(white(?x+1,?y+1)))
:effect (open(?x,?y) white(?x+1,?y+1))
\end{lstlisting}

\begin{lstlisting}[caption={Breakthrough: Example Problem}, label={lst:breakthroughproblem},language=stripspddl,mathescape]
#boardsize
2 4
#init
(black(1,4) black(2,4) black(1,3) black(2,3)
white(1,1) white(2,1) white(1,2) white(2,2))
#depth
13
#blackgoal
(black(?x,ymin))
#whitegoal
(white(?x,ymax))
\end{lstlisting}

\begin{lstlisting}[caption={KnightThrough: Domain Snippets}, label={lst:ktdomainsnippets},language=stripspddl,mathescape]
#blackactions
:action L1
:parameters (?x, ?y)
:precondition (black(?x,?y) NOT(black(?x+1,?y+2)))
:effect (open(?x,?y) black(?x+1,?y+2))
:action L2
:parameters (?x,?y)
:precondition (black(?x,?y) NOT(black(?x+1,?y-2)))
:effect (open(?x,?y) black(?x+1,?y-2))
...
#whiteactions
:action L1
:parameters (?x,?y)
:precondition (white(?x,?y) NOT(white(?x+1,?y+2)))
:effect (open(?x,?y) white(?x+1,?y+2))
:action L2
:parameters (?x,?y)
:precondition (white(?x,?y) NOT(white(?x+1,?y-2)))
:effect (open(?x,?y) white(?x+1,?y-2))
...
\end{lstlisting}

\begin{lstlisting}[caption={KnightThrough: Example Problem}, label={lst:ktproblem},language=stripspddl,mathescape]
#boardsize
3 4
#init
(black(1,4) black(2,4) black(3,4) black(1,3) black(2,3) black(3,3)
white(1,1) white(2,1) white(3,1) white(1,2) white(2,2) white(3,2))
#depth
1
#blackgoal
(black(?x,ymin))
#whitegoal
(white(?x,ymax))
\end{lstlisting}

\begin{lstlisting}[caption={Evader-Pursuer: Domain Snippets}, label={lst:evdomainsnippets},language=stripspddl,mathescape]
#blackactions
...
:action right-diagonal-down
:parameters (?x, ?y)
:precondition (black(?x,?y) open(?x+1,?y+1))
:effect (open(?x,?y) black(?x+1,?y+1))
:action stay
:parameters (?x, ?y)
:precondition (black(?x,?y))
:effect (black(?x,?y))
#whiteactions
...
:action left
:parameters (?x, ?y)
:precondition (white(?x,?y))
:effect (open(?x,?y) white(?x-1,?y))
:action stay
:parameters (?x, ?y)
:precondition (white(?x,?y))
:effect (white(?x,?y))
\end{lstlisting}

\begin{lstlisting}[caption={Evader-Pursuer: Example Problem}, label={lst:evproblem},language=stripspddl,mathescape]
#boardsize
8 8
#init
(black(8,1) white(2,3))
#depth
11
#blackgoal
(black(xmin,ymin))
#whitegoal
(white(xmin,ymin))
\end{lstlisting}

\begin{lstlisting}[caption={Domineering: Domain}, label={lst:domineeringdomain},language=stripspddl,mathescape]
#blackactions
:action vertical
:parameters (?x, ?y)
:precondition (open(?x,?y) open(?x,?y+1))
:effect (black(?x,?y) black(?x,?y+1))
#whiteactions
:action horizontal
:parameters (?x, ?y)
:precondition (open(?x,?y) open(?x+1,?y))
:effect (white(?x,?y) white(?x+1,?y))
\end{lstlisting}

\begin{lstlisting}[caption={Domineering: Example Problem}, label={lst:domineeringproblem},language=stripspddl,mathescape]
#boardsize
6 6
#init
#depth
11
#blackgoal
#whitegoal
\end{lstlisting}

\end{document}